**(Pre-print) AAS 16-510**

# PRELIMINARIES OF A SPACE SITUATIONAL AWARENESS ONTOLOGY

**Robert J. Rovetto[*], T.S. Kelso[†]**

Space situational awareness (SSA) is vital for international safety and security, and the future of space travel. By improving SSA data-sharing we improve global SSA. Computational ontology may provide one means toward that goal. This paper develops the ontology of the SSA domain and takes steps in the creation of the space situational awareness ontology. Ontology objectives, requirements and desiderata are outlined; and both the SSA domain and the discipline of ontology are described. The purposes of the ontology include: exploring the potential for ontology development and engineering to (i) represent SSA data, general domain knowledge, objects and relationships (ii) annotate and express the meaning of that data, and (iii) foster SSA data-exchange and integration among SSA actors, orbital debris databases, space object catalogs and other SSA data repositories. By improving SSA via data- and knowledge-sharing, we can (iv) expand our scientific knowledge of the space environment, (v) advance our capacity for planetary defense from near-Earth objects, and (vi) ensure the future of safe space flight for generations to come.

## INTRODUCTION

*Space situational awareness* (SSA) is vital for international safety and security. Of paramount importance is the early detection of potential hazards to astronauts, space-borne assets, and our terrestrial home. Improving the state of SSA is a global necessity, one that requires international cooperation, ever-advancing sensor networks, and analyzing and sharing SSA data. Achieving an ideal state of SSA is arguably to achieve actionable, real-time, predictive awareness of the space environment. To move toward such a state, we need to improve our data-sharing capabilities. This paper focuses on one research area to achieve this: ontology development.

*Ontology* is the general study of reality or any domain of interest. It is the study of the sorts of objects and their (inter)relationships in a given domain. Formal methods and ontological categories in this philosophical discipline are often applied to computer and information science. The products are *computational ontologies*, computable artifacts representing the individuals, kinds and relations of a domain. They are formal theories representing domain objects and expressing domain knowledge in a computable format. These artifacts are used, in part, to annotate data, and foster data-exchange, interoperability, and communicate a conceptualization. Computational *ontology development and engineering*, then, is the process by which these ontologies are designed, developed and implemented.

---

[*] Corresponding author. Email: rrovetto@terpalum.umd.edu ; ontologos@yahoo.com. Ontologist, philosopher. NY, USA. University of Maryland Alum 2007; University at Buffalo SUNY Alum 2011; APUS/AMU, space studies
[†] Senior Research Astrodynamicist, Center for Space Standards and Innovation, Analytical Graphics Inc. Email: tskelso@agi.com

In what follows, preliminaries of a space situational awareness ontology are presented. This paper develops the ontology of the SSA domain and takes steps in the creation of the Space Situational Awareness domain Ontology (SSAO). It outlines ontology objectives, requirements and desiderata; and describes both the domain and the discipline of ontology. The goals of an SSAO are at least twofold. First, to formally represent the domain in a scientifically useful and accurate manner: its objects, the space environment, how they interact, the patterns thereof, and the processes by which we maintain awareness of these objects. The second goal is to improve global SSA by fostering data- and knowledge-exchange in the space community; and advance our scientific knowledge in the process. An assumption, then, is that SSA data-sharing will, indeed, improve SSA, i.e., it will improve space safety, and our capacity for planetary defense. The intended use of the ontological system is to help solve problems in the space domain, such as preventing satellite collisions; minimizing orbital debris formation; and improving early-detection of potentially hazardous near-Earth or deep-space objects.

There is little[*] ontology development efforts in the SSA or broader space domain as compared to other disciplines. In this respect, we offer novel concepts. This paper follows ideas introduced in Rovetto(2015)[1], which presented a project concept conceived with the discovery that the orbital debris problem may benefit from more data exchange and integration[†]. The overall idea is motivated from both a passion and intellectual fascination for astronautics and astrodynamics, and a desire to help ensure safe spaceflight by, in part, solving the orbital debris problem. Ontology development, both philosophical and computational, is a research field that has the potential to improve SSA, and thereby help prevent and solve space domain problems.

If we are to achieve real-time responses to rapidly changing orbital events and potential space environment threats, SSA data must be *dynamically* updated and available in real-time. We therefore state this caveat. Given the current state of the art in computing, there is the possibility that ontologies may slow computational processes when reasoning over large data-sets in real time[‡]. Ontologies should, therefore, be used to the extent that they (a) do not hinder space safety and SSA[§], and (b) contribute to achieving the above goals. In short, the priority—improving space safety and planetary defense via greater global SSA—must guide research tracks. This paper takes steps in one track: *ontology for the SSA domain*.

The paper is divided thusly: the domain to be ontologically characterized is first described, followed by a summary of the discipline of ontology. Desiderata for an SSA domain ontology is listed, the SSAO is introduced with part of its taxonomy, and an example first-order formalization is presented. Steps/tasks in the development process is marked with '(S#)' and suggested guide-lines by '(R#)'. Italics or bold marks key terms. Bold and camel-cased terms are unary category terms. Italicized and camel-cased terms are relation terms.

---

[*] For two early efforts, see [17], and [18]. The former is a schema, not an ontology, but has many essential terms for SSA taxonomies. The latter has terms from different scientific disciplines relevant to SSA.
[†] Thanks to David Vallado (Analytical Graphics Inc.) for making the need for data exchange/integration known to the corresponding author (via a conference presentation in Boulder, CO, 2011).
[‡] Personal communication with Lowell Vizenor
[§] E.g. if unable to handle real-time dynamically changing data, or the physics- and mathematics-intensive aspects pects of the domain; large automated inference times; etc. The utility of *dynamic ontologies* should therefore be investigated.

## THE SPACE SITUATIONAL AWARENESS DOMAIN

The universe of discourse to be expressed in an ontological framework is the SSA domain. *Space situational awareness* is *situational awareness of the orbital, near-Earth and deep-space environments*. It includes the processes by which we achieve that awareness, such as observation, detection, identification, tracking, and prediction/propagation of space objects, their orbits and trajectories; as well as phenomena in the space environment. Elsewhere SSA has been defined as:

- "the ability to view, understand and predict the physical location of natural and manmade objects in orbit around the Earth, with the objective of avoiding collisions"[2]

- "understanding and maintaining awareness of the Earth orbital population, the space environment, and possible threats." [3]

- "[…] the ability to accurately characterize the space environment and activities in space." [4]

The last quotation captures a central purpose of SSA ontology: to formally characterize the space environment and activities, events and processes thereof. A broader context of related space entities is therefore associated with SSA. It will be helpful to either identify or delimit subdomains in order to better manage the subject matter. Partially overlapping divisions of domain content will facilitate SSAO development.

### SSA Activities and Goals

Table 1 lists SSA activities and areas from European and United States perspectives. According to [4, p.2], the goals of SSA from the perspective of the latter include "characterising, as completely as possible, the space capabilities operating within the terrestrial and space environments".

**Table 1. SSA Sub-divisions according to EU and USA.**

| European Space Situational Awareness Program [16] | United States |
|---|---|
| - Space surveillance and tracking<br>- Space weather effects<br>- Near-Earth objects | - Intelligence<br>- Surveillance<br>- Reconnaissance<br>- Environmental Monitoring<br>- Command and Contro |

These activities involve: observing natural and artificial objects in the space environment, reasoning over accumulated data, predicting future space object motion, and taking actions to avoid hazardous situations. Together they form a SSA whole whose purpose is to ensure safe space and terrestrial activity. To structure the domain, we assert three naturally overlapping benefit- and goal-based categories are as follows.

I Planetary Defense
- Orbital awareness (orbital debris, active satellites, etc.)
- Near-Earth awareness (e.g. asteroids, comets)
- Deep-space awareness (comets, interstellar phenomena, etc.)

- Space weather awareness and forecasting (solar activity, etc.)
II   Protection of orbital *in situ* persons and space assets (communications satellites, stations)
III  Spaceflight safety, Space traffic management

More specific reasons for SSA, drawn largely from [5] and [6], are here organized into additional activity-based (processual) categories:

**PRODUCING**:   Running catalogs of space objects

**PREDICTING**:  - Collisions in orbit
- Calculating the risk to spacecraft due to environmental threats
- Chart the present position of orbital objects and plot their anticipated orbital paths.
- Atmospheric re-entry of space objects; When and where a decaying space object will re-enter the Earth's atmosphere.

**PREVENTING**:  - Collisions on orbit
- A returning space object, which to radar looks like a missile, from triggering a false alarm in missile-attack warning sensors

**DETECTING**:   - Hazards to spacecraft
- Malfunctions
- New space objects

**IDENTIFYING**: Which country owns a re-entering space object

**MONITORING**:  Behavior of spacecraft, e.g. changes in altitude, position, etc.

**DIAGNOSING**:  Spacecraft failures and malfunctions

In short, space situational awareness includes at least:

(A) **Observation** of the space environment,
(B) **Identification** and **Tracking** of **space objects** in that environment,
(C) **Accumulation** and **Analysis of Data**, and
(D) **Knowledge discovery** that ideally is **actionable**

Ground- and space-based sensor networks are used to observe the orbital and near-Earth environments. Some SSA networks include the following. For more details on sensors see [7] and [8].

- International Scientific Optical Network [9]
- Canadian Space Surveillance System [10]
- Space Surveillance Network (SSN) [6]
- Russian Space Surveillance System
- Chinese Space Surveillance System
- Space Data Association [11]

To better achieve the above goals and improve global space safety, sensor networks in conjunction with satellite operators around the globe must share SSA data. One potential challenge is that each space actor may use different data formats; have unique database terms referring to the same space object; and their databases (e.g. space object catalogs) may be entirely isolated from one another. Toward resolving these challenges, **ontologies** offer structured, sharable, interoperable and computable taxonomies that have a formal semantics. They formally represent common and tacit domain knowledge shared by SSA communities as well instance data about the respective domain objects. This allows semantic interoperability among SSA actors.

Space communities around the globe have overlapping knowledge: the science and engineering of astrodynamics, astronomy, satellite operations, aerospace engineering, etc. SSAO formally represents some of this general scientific knowledge, and the entities it is about, in one or more potentially interconnected and modular ontologies. Given the wide and interdisciplinary scope of SSA, an SSAO is more accurately an SSAO suite that includes specific domain ontologies. These computable terminological systems contain explicitly defined classes that can be mapped to one another, and that can annotate or subsume terms from SSA databases, affording interoperability among SSA information systems. An SSAO ontology thereby has the potential to improved SSA for the respective data-sharing space actors. It also may help glean insights into novel *astrodynamic standards* by, in part, putting forth a community SSA vocabulary.

## ONTOLOGY AND COMPUTATIONAL ONTOLOGIES

Ontology in computer science circles is distinguished, but related to, philosophical ontology, the latter of which is general study and characterization of actual and potential existence. A philosophical ontology, then, is a theory of the kinds of entities that (are held to) can or do exist and their interrelationships. **Ontology/ontological engineering** [12] has been described as:

> "the set of activities that concern the ontology development process, the ontology life cycle, the methods and methodologies for building ontologies, and the tool suites and languages that support them" [13].

This involves the specification of a computable terminology with a formal semantics: a computational ontology. The meaning of the terms composing the taxonomy is expressed in natural and artificial languages. Good ontology practice calls for one meaning per term to avoid ambiguity and confusion. **Computational ontologies** (also called *information* or *applied ontologies*), then, are computable systems of terms whose intended meanings are represented in an ontology language. As such:

> "[t]he ontology engineer analyzes relevant entities and organizes them into concepts [classes] and relations, being represented, respectively, by unary and binary predicates. The backbone of an ontology consists of a generalization / specialization hierarchy of concepts, i.e., a taxonomy." [12].

Organizing relations, such as *class subsumption* (*is a*), are used to organize the terms. The *is a* relation can be defined as: some class A is a subclass of class B if and only if A inherits all properties of B. Partonomies are taxonomies describing the partonomic relationship between entities, and uses one or more *parthood* relation. For example, *part of* is often defined according to General Extensional Mereology.

In both philosophical and computational ontology, **categories** (types, universals, classes) are often distinguished from their **instances** (tokens, particulars, members, individuals). They are relatable with an *instantiation* (*instance_of*) relation.

Each class in the ontology should be given a definition, save primitives[*]. Primitive terms should be given clarifying comments to aid the ontology user in grasping the general sense of the term. Definitions are subject to revision over time as scientific and domain knowledge changes. Definitions often take the form of asserting *necessary and sufficient conditions*, which helps automated reasoning, but other sorts of definition are possible. Natural language definitions convey the meaning of terms to human users, including ontology curators and developers. Artificial languages, such as knowledge representation or ontology languages, are used to make the terms computable. Logical formalisms such as first-order predicate calculus are used to help create formal definitions. Thus, two central steps in the ontology development process are forming a vocabulary of terms within the scope of the domain, and defining them. Ontology terms are used to **annotate** instance data (data about individuals in the world, e.g., the Hubble Space Telescope). Types of SSA instance data includes observational data (e.g. infrared, optical data), and data about the orbital parameters of some individual satellite.

First-order, modal and higher-order logics are used, in part, to test for correct inferences in the less expressive computational implementation languages (artificial languages) such as Common Logic (CLIF)[14], and OWL[15]. Any given formalism—from modal logics to implementation languages—has limitations, e.g., limited expressivity. There are also different ways to symbolically represent and computationally implement a given ontological theory. In any case, the implementation language should attempt to capture the full intended meaning (at the conceptual and natural language levels) of terms. Where a mismatch between intended meaning and the implementation exists, it should be explicitly stated in documentation and ontology files to avoid misinterpretations (R1). Table 2 lists some general functions and goals of ontologies.

**Table 2**. **Goals of computational ontologies**

| Computational Goals | Conceptual Goals/Benefits |
|---|---|
| Annotation | Semantic clarity, Explaining the meaning of domain terms and data |
| Automated Inference/Reasoning | |
| Data sharing, Exchange, Integration | Conceptual and philosophical explication |
| Data representation | Presenting a shared conceptualization |
| Interoperability | Knowledge representation and Reuse |

The applied ontology development process should include the *open world assumption* (R2) and must be *subject to revision and correction* (R3) over time. It is an iterative process involving formal and concept(ual) analysis; development; implementation; validation and testing. Software development methodologies may be adopted. Philosophical ontology informs this process with formal distinctions and tools, just as scientific knowledge inform the philosophical descriptions of the domain.

Computational ontologies may draw upon philosophical ontology by employing highly general **distinctions** and **ontological categories,** such as the following.

| | | |
|---|---|---|
| Space | Concrete Particular | Identity |
| Time | Abstract Particular | Persistence |
| Space-Time | Entity | Modality |
| Event | Object | Continuant / Endurant |

---

[*] Primitive terms are those that are undefined within the system.

|  |  |  |
|---|---|---|
| System | Process | Occurrent / Perdurant |
| State | Property-bearer | Universal vs. Particular |
| Function | Property | |

These categories, which are given symbolic definitions in formal ontology, are related to one another with **formal** (**domain-neutral**) **ontological relations** such as the following.

|  |  |  |
|---|---|---|
| Dependence | Causation | Parthood |
| Inherence | Participation | Composition |
| Instantiation | Connection | Constitution |

Various sub-relations of Dependence (and other relations) can more specifically characterize the actual physical, material and relational dependencies among the entities in the SSA domain. Parthood and composition are mereological relations, where *mereology* (and mereotopology) is the general study of the relationships between parts and their wholes (and connectedness). Additional tools for ontological analysis include formal theories of **unity**, and **identity**.

Note that there are different accounts of each of the above concepts. There is arguably no universal agreement as to their ontological status, e.g., as to whether causation is indeed a relation. The SSAO, like other domain ontologies, may therefore: (a) assert its own treatment on the respective concept, (b) adopt existing ones, or (c) adopt an ontology methodology that does not commit to such philosophical distinctions.

Finally, ontological inquiry into SSA (specifically astrodynamics) has a large **epistemological** and **modal** component. That is, SSA involves knowledge of the *present* situation (detecting an existing space object), current events and processes (detection of collision events, ongoing spacecraft operations, maneuvers, etc.), physical states and properties (shape, mass, the Keplerian orbital parameters), and very importantly *predictive* (or *future*) knowledge. The latter involves extrapolating possibilities, such as potential collisions, orbital paths, etc. It is therefore critical for a SSAO to capture the prediction, propagation, and modality aspects of the domain (R3).

**APPLYING ONTOLOGY TO THE SPACE SITUATIONAL AWARENESS DOMAIN**

There are different ontology development approaches [20], but developing a cogent and working Space Situational Awareness Domain Ontology includes at least steps S1 through S5.

(S1) **Identify**: domain problems to solve, goals, requirements, and questions

(S2) **Domain research**: reference documents, domain-experts, domain data & databases

(S3) **Demarcation** of sub-domains for better content management (context-specific)

(S4) **Vocabulary/Terminology:** List domain-specific terms to be formed into a taxonomy. Concept(ual development

(S5) **Definitions** of terms from S4 using natural and artificial language definitions, including formal rules and logical axioms to capture domain knowledge.

General goals, S1, include SSA data-exchange among civil, federal and military SSA actors. A more specific goal is the **sharing of unmediated data** between interested space actors in order to minimize time between observations. This will lower response time to potential or imminent threats to space assets. If international SSA communities use different data formats, then ontology offers an avenue toward interoperability.

S2 includes consulting domain literature, research groups, individuals, space object catalogs and databases, space agencies, SSA sensor networks, and so on. It is essential for a variety of practicing subject-matter professionals with different viewpoints and ideas to be involved. Domain professionals help explain, verify and correct domain knowledge expressed by the formal ontological representations of ontology developers and curators. They therefore help ensure faithfulness to domain, but also stand to gain insights from formal and philosophical ontologists. Ontology developers and curators will ideally be domain experts (or vice versa). Toward this, educational courses in SSA-related topics for ontologists should be provided (R4). If an ontological approach according to which existing ontologies are reused, both domain-experts and ontologists should evaluate all ontology resources [21].

One function of an SSAO is to symbolically represent and computationally implement SSA knowledge. Toward S4 and S5, we form a SSA **taxonomy**, assert the interrelationships between terms (mirroring real-world relations among their referents), and structure the terms into a hierarchy using the class-subsumption relation (or otherwise). Class terms may be organized along the dimensions of SSA subareas and activities discussed in section 2, or along other dimensions and domain sub-groupings. A SSAO should have domain-specific category terms for some or all of the following entities, grouped into categories marked by "(T#)":

**(T1)** **SPACE OBSERVATIONS** (an observation as distinct from the observed)

**(T2)** **SPACE OBJECTS & PHENOMENA** being observed
- *Classify* space objects and phenomena: Satellites, Spacecraft, Orbital Debris, Asteroids, Space weather phenomena, etc.

**(T3)** **OBSERVATION PROCESSES** engaged by space operators, astrodynamicists, astonomers, sensors, etc.
- Detection (e.g. Detection Event)
- Identification
- Tracking
- Propagation

**(T4)** **DATA** (from observations) representing or measuring the observed objects (or some property thereof)[*]

**(T5)** **SENSORS** that gather data from observations, and that engage in observations

Each of these potentially constitutes the subject matter of a distinct and modular, yet interoperable, ontology (or a portion thereof) within a global **SSA Domain Ontology Architecture**. For example, an **SSAO suite** can consist of a(n):

- **Ontology of Space Observation Processes and Procedures**
- **Space / Satellite Operations Ontology**

---

[*] It will help to be clear on distinctions between data, observations, and what data is about or what it refers to (if anything). This will help avoid category mistakes and misrepresentations.

- **Spacecraft and Sensor Ontologies** for space assets, sensors, etc.
- **Space Object Ontology**
- **Orbital Event and Process Ontology**
- **SSA Data Ontology** (representing data formats)

… and so on.

Although the domain to be ontologically represented is broader than a literal or narrow reading of 'SSA', an alternative would be to focus the scope of an SSA ontology (by the same name) to space awareness activities and objects (i.e., observation, detection, tracking, prediction activities; communications; sensor-networks, etc.), leaving other entities, such as space environment phenomena & objects, and orbital dynamics knowledge, to be represented in similarly focused ontologies, or in one or more broader space ontologies, all of which can be interconnected.

Any ontology will have one or more ontology files, implemented in a computable language such as Common Logic (CLIF)[14] or OWL[15], the former of which is more expressive and recommended between the two. Ontology class definitions are formalized in such ontology languages.

To represent the shared general scientific knowledge relevant for SSA activities, **modular scientific domain ontologies** for each discipline are appropriate. **Astrodynamics**, and the physical principles therein, for instance, is a necessary subject matter to capture. Awareness of space debris—and with it **conjunction analysis**[*]—is a major part of SSA. Following [1], an Orbital Debris Ontology serves to enable space debris data-sharing and thereby improve spaceflight safety and SSA. If the astrodynamics and orbital debris domains are not large enough to form individual ontologies unto themselves, then the respective classes shall be part of the class hierarchy of a SSAO.

The international SSA community utilizes similar concepts and terms, largely in virtue of this common scientific knowledge. The domain is also interdisciplinary, using concepts from astrodynamics, general physics, and astronomy. Some terms will more precisely belong to a specific scientific, operational, or engineering discipline. Each discipline may have a corresponding domain-ontology. In any case, a degree of arbitrariness will go into grouping the terms and demarcating the knowledge and domain to be represented by each ontology. Existing domain ontology resources such as [18] or [19], where physical and astronomical terms abound, may make this process more efficient if the relevant class terms can be reused.

For example, although the class **Asteroid** would be accurately placed as part of the taxonomy of an Astronomy Ontology, such as [19] (as type of **Astronomical Object**), it may be formally represented and categorized differently by distinct databases or ontologies. Alternative placements are in a Near-Earth Object Ontology, a Space Environment Ontology, Space Weather Ontology, a Space Object Ontology (as a type of Space Object), and so on and so for. Various classification schemes are possible, and occasionally arbitrary, but consulting scientific knowledge of the entity in combination with formal ontological tools, should yield a scientifically accurate classification. The physical, intrinsic and essential properties of these space entities may inform their taxonomic and domain placement. The SSAO could then import the class from its respective ontology, relating it to others. This action is accomplished using ontology/taxonomy editor applications.

---

[*] Predicting potential collision events, i.e. future possibilities.

Table 3 (as well as 4 and 5) presents a sample of relevant class terms for the SSAO, and any similar ontology seeing to represent the given domain. The meaning of most terms is straightforward, but asserting definitions is necessary. Some terms are commonly found in the space community, others are offered as novel additions. Those commonly found can be drawn from existing SSA and other space terminology resources, e.g., space object catalogs, SSA databases, and space agency and scientific literature. Notice that the terms are for different sorts of entity (or can be classified as such): natural celestial bodies (e.g., asteroids); space artifacts (e.g., spacecraft); information and data objects (labels; names; data formats, e.g., TLE); and properties (physical, geometric, social). Also note that terms such as Astronomical Body/Object need not be asserted by the SSAO, but can be asserted by and imported from another space domain ontology, just as with the asteroid example.

**Table 3. General terms for a Space Situational Awareness Ontology.**

| | | | |
|---|---|---|---|
| Spacecraft, Space Vehicle | Orbit | Satellite Number | Orbital Conjunction |
| (Artificial) Satellite | Orbital Element / Parameter | Satellite Catalog Number | Orbital Collision Event |
| Communications Satellite | | | Collision Avoidance Maneuver |
| Orbital Debris | Orbital Period | COSPAR ID | Astrodynamic Process |
| Sensor | Orbital Inclination | NORAD ID | Space Object Tracking Process |
| Space-based Sensor | Eccentricity | Operator | Space Object Detection Event |
| Ground-based Sensor | Epoch | Owner | Space Weather Event |
| Optical Telescope | Perigee, Apogee | Launch Date | Space Operations |
| Astronomical Body | Right-Ascension of the Ascending Node | Two-line Element Set (TLE) | Space Contact |
| Space Object | | | |

For each class of entity we should (S6) determine their:

- Properties, features, or attributes
- Identity and unity conditions/criteria
- Dependencies and interrelationships
- Parent categories

**Properties** of objects are often philosophically described as **Dependent Entities**. *Identity* and *unity* conditions are typically considered necessary conditions that indicate the identity or equality of some entity. *Identity conditions* are that without which an entity of a given sort would not be of that sort. Dependencies are those states of affairs and entities (objects, relations, processes, properties, etc.) that the entity in question relies on, existentially or otherwise. Parent categories indicate the minimum properties characterizing a child category. For example, a Telecommunications Satellite category is a sub-category of Communication Satellite and Artificial Satellite (at a higher level of abstraction), the former inheriting the properties of the latter two. Examples of these properties include having a particular function/purpose and having been made by persons.

In other words, conduct an ontological analysis to define terms, capture the intended meaning, and give a precise formal semantics. Table 4 lists some specific property and relation terms of interest. Indentation indicates class subsumption. Relations are represented as n-ary (at least binary) predicates.

Table 4: Property and Relation Terms.

| Property (unary predicate) | Relation (n-ary, n≥2) |
|---|---|
| Mass | Has Orbit |
| Material Composition | Has Orbital Element |
| Shape | Has Inclination, Has Eccentricity, … (through the orbital elements) |
| Radar Cross-section | Has Cross-section |
| Function / Purpose  Design Function | Has Property, Has Function |
| Albedo | Has Status (e.g., Operational, Inactive, Defunct, Abandoned) |

Table 5 presents the domain and range of some binary predicates (expressing binary relations). Each row should be read from left to right as a formal statement in the ontology, e.g., **Satellite *has_orbit* Orbit**. This reflects the simple form of Subject-Predicate-Object similar to RDF triples (Resource Description Framework format), but more expressive and complex statements are possible as when working with first- and higher-order logics (see final section). Entities that have numerical values, such as the orbital parameters, can also be modeled as a predicate taking that value. Ontology editors, such as Protégé, use what are called 'datatype property' to do so: the individual orbit (or satellite) term would be linked with a decimal value via a has_Orbital_Eccentricity datatype property. According to that model, an Eccentricity class may be omitted. However, it remains to be seen whether this is the best approach, and a more expressive ontological representation would arguably retain the class.

Table 5: Relations with candidate domain and range.

| Domain | Relation | Range |
|---|---|---|
| Artificial Satellite | *Has_Status* | Satellite_Operational_Status (*Values*: Operational, Active, Inactive, Defunct, Abandoned, etc.) |
| Satellite | *Has_Orbit* | Orbit |
| Orbit | *Has_Orbital_Inclination* | Inclination (Example value: 60°) |

Ontology classes will **annotate** instance data housed in SSA databases, and should explicitly and clearly communicate what the data is about. This is a basic goal of ontologies. **Space object catalogs**—data repositories of instance data about actual objects in Earth orbit—are therefore to be annotated with the relevant space object categories: Spacecraft, Space Vehicle, GPS Satellite, Active Satellite, Orbit, Rocket Body, Orbital Debris, Space Telescope, Space-based Sensor, Space Station, etc. For example: **Hubble_Space_Telescope *is_instance_Of* Space-Borne_Telescope**.

An SSAO terminology will therefore provide general class terms common to the subject matter shared by each space actor and their databases. Each space actor annotates their database terms with classes from the SSAO (or SSA domain ontology suite, consisting of more than one interconnected space ontology), ideally facilitating data-exchange. Furthermore, there are different ontology methods and architectures to assess. For example, each space actor can create their own local ontology for their SSA data. These local ontologies can then be interconnected in a number of ways: by using the SSAO (or suite) to subsume their own classes; by performing mappings of classes between each local ontology, asserting some classes equivalent/synonymous; etc. Ontologies can therefore serve to relate and map each space actor terminology to one another.

Research into and application of various ontology development approaches should serve to help solve space domain and space data challenges, e.g., orbital debris remediation, space data exchange and integration, as well as help cultivate safe space flight and development by offering both data- and knowledge-modeling capabilities and a means to facilitate SSA data-sharing.

Figure 1 presents preliminaries of the working taxonomy of the SSA domain ontology (subject to revision). At the time of, and prior to, this publication, steps in the ongoing evolution of the ontology file[*] and an SSA Vocabulary/Data Dictionary were commenced by the corresponding author.

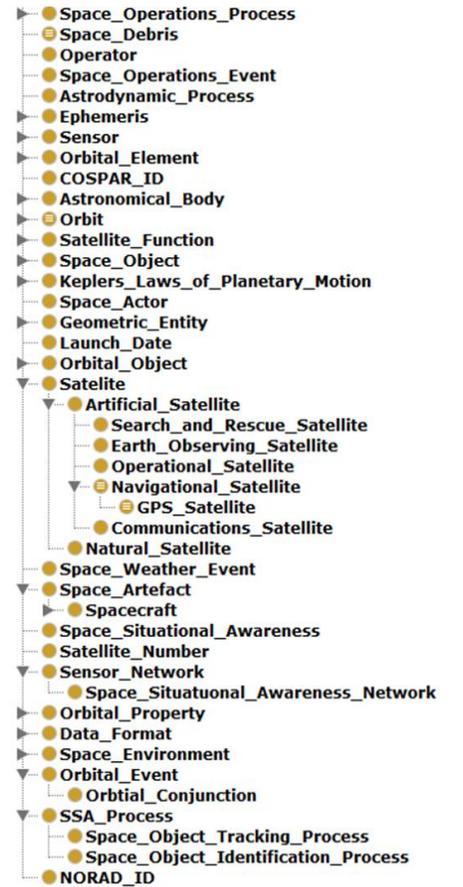

**Figure 1: A working taxonomy for the SSAO, displayed in Protégé.**

**A FORMALIZED SSA SCENARIO**

To visually express the idea of a space situational awareness ontology, Figure 2 is a diagram of SSA categories and relations. It depicts a fictional scenario in which a particular satellite is tracked by a sensor that is part of a specific SSA network. The top half above the dotted line represents class-level terms (expressing general knowledge). The lower half represents instances of those classes. Red arrows represent the instantiation relation between the general (class) and the particular (individual). A generic *Part Of* relation is used, but undefined.

---

[*] The file is presently located on Github. https://github.com/rrovetto/space-situational-awareness-domain-ontology

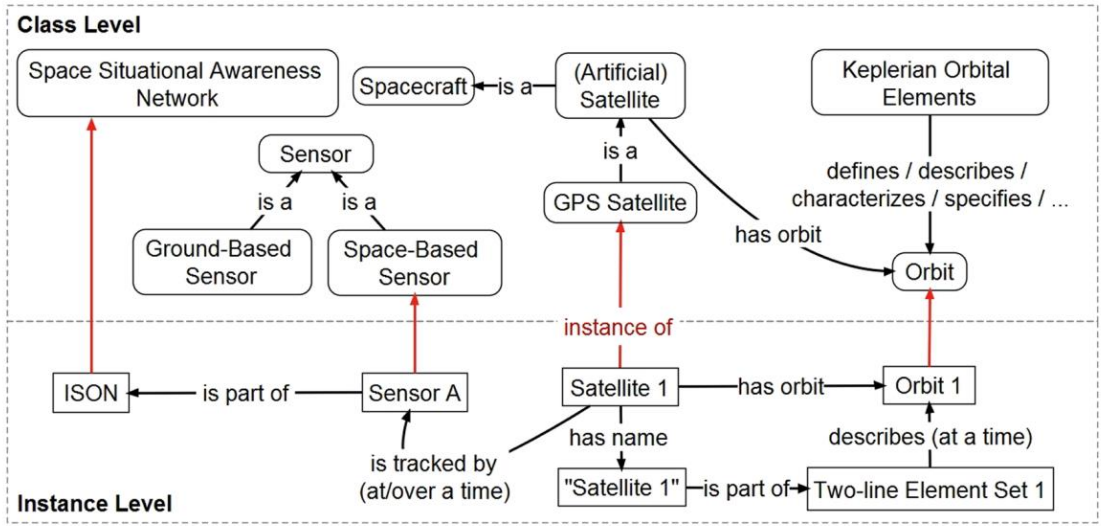

**Figure 2: An ontological diagram of some SSA-relevant categories. Rounded rectangles signify classes, rectangles instances, and arrows with italicized text represent relations. Red arrows crossing the dotted line mark the instantiation relation between individuals and their general category or class.**

To understand the level of detail that goes into the requisite ontological analysis, note some limitations of Figure 2. The classes and relations are not defined here, but when doing so we must *reference times or temporal intervals*. For example, an instance-level ***Tracked By*** relation may be represented by a ternary predicate relating **Satellite**, **Sensor** and time classes. The scenario assumes the Global Positioning Satellites are, in fact, in orbital motion, an assumption consistent with an intuitive conception of artificial satellites as an artifact in orbital motion about another body. Given Figure 2, being in orbital motion is a property (or state) that needs to be explicitly formalized. By contrast, if we define **Artificial Satellite** as an artifact whose **Function** is to orbit the Earth, then the class-level ***Has_Orbit*** relation should be omitted since it would not hold atemporally. The reason is that prior to orbit-insertion, any given artificial satellite may be resting on the surface of Earth. Finally, some space-based sensors such as the Hubble Telescope are satellites in the sense of being orbiting artifacts. These and other considerations must be taken into account to refine SSAO and ensure coherence and clarity.

A definition of the class, GPS_Satellite, that is computable when part of a coherent ontology is as follows.

**GPS_Satellite** =def. An **Artificial_Satellite** that is *part_of* the **Global_Positioning_System**

In other words, the definiendum is a subclass of **Artificial_Satellite** with differentiating properties or relationships of *being part of* the GPS.

First-order predicate logic (FOL) axioms for Figure 2, along with their natural language (NL) reading (in italics), are as follows. Standard FOL constants and connectives are used. 't' denotes temporal instants.

∀ ("For all" Universal quantifier)     → ("if then"'/ implication)

∃ ("There exists"/Existential quantifier)     ∧ ("and"/conjunction)

Is_a(Space-Based_Sensor, Sensor) (A1a)
*All space-based sensors are sensors.*

∀x[ instance_of(x, Space-Based_Sensor) → instance_of(x, Sensor)] (A1b)
*For all x, if x is an instance of (the class) Space-Based_Sensor,
then x is an instance of Sensor.*

Is_a(GPS_Satellite, Artificial_Satellite) (A2a)
*All GPS Satellites are Artificial Satellites.*

∀x[instance_of (x, GPS_Satellite) → instance_of (x, Artificial_Satellite)] (A2b)
*For all x, if x is an instance of GPS_Satellite, then x is an instance of Artificial_Satellite.*

instance_of(Sensor A, Space-Based_Sensor) (A3)
*Sensor A is an instance of Space-Based_Sensor.*

From A1 and A3, Sensor A is also a(n indirect) *instance_of* Sensor. An automated reasoner will make this inference if the classes and axioms are defined and specified properly. If multiple inheritance is desired, then assert an ***is a*** relation between Space-Based Sensor and Spacecraft as well. This and other considerations depend on how we define the classes and what distinctions we adopt, e.g. Artificial-Natural, etc. A more complicated expression is (A4) and (A5).

Every satellite tracked by Sensor A has some particular Two-Line Element set (A4)
(which describes the orbit of the satellite).

∀x [ instance_of(x, Satellite, t) ∧ is_tracked_by(x, SensorA, t) →
 ∃y,z,t [ instance_of(y, Two-Line_Element_Set)
 ∧ instance_of(z, Orbit)
 ∧ describes(y, z, t)]]

*For all x, if x is an instance of Satellite at time t, and x is tracked by SensorA at time t, then there exists a y, a z and a time t such that y is an instance of Two-Line_Element_Set and z is an instance of Orbit and y describes z at time t.*

The alternative formalization, A5, removes Orbit classes, and asserts a relation such as ***Describes_orbit_of***.

∀x [ instance_of(x, Satellite, t) ∧ is_tracked_by(x, Sensor A, t) → (A5)
 ∃y,z,t[instance_of(y, Two-Line_Element_Set) ∧ describes_orbit_of(y, x, t)]]

To formally express the orbital parameters expressed in a TLE, relate the orbit with each parameter, e.g., has_orbital_inclination(Orbit1, 60°).

*\* \* \**

This example concludes the paper by demonstrating a sample of the formalization required. Further work is necessary, but this is part and parcel of what goes into the formal and applied ontology process for the space situational awareness domain.

## CONCLUSION

This paper has presented the foundations of space situational awareness ontology, outlining requirements and desiderata for formal ontologies and space taxonomies for the SSA domain. The discipline of ontology (philosophical and computational) was described; the SSA domain to be ontologically represented was summarized and demarcated, and some key class terms identified. Early stages of the applied ontology, the Space Situational Awareness Domain Ontology (SSAO), was introduced via part of its working taxonomy. Finally, a sample first-order formalization was presented.

The goals of an SSAO or ontology suite are to: provide formal and computable representations of general scientific knowledge (e.g., astrodynamics), domain objects (satellites, orbital debris, etc.) and inter-relations; annotate SSA instance data; and foster space data-sharing. Space object catalogs containing satellite observational data can be annotated with corresponding ontology classes to afford data-exchange and semantic interoperability. The overarching purpose of these goals is to improve peaceful SSA and spaceflight safety for the global space community. SSA is a global necessity that thereby offers us an opportunity for international cooperation among space actors in all sectors: government, private, academia.

## REFERENCES

[1] Rovetto, Robert J. (2015)"An Ontological Architecture for Orbital Debris Data." Earth Science Informatics. 9(1), 67-82. Springer Berlin Heidelberg. URL=
http://link.springer.com/article/10.1007/s12145-015-0233-3
DOI: 10.1007/s12145-015-0233-3.

[2] "Space Foundation" http://www.spacefoundation.org/programs/public-policy-and-government-affairs/introduction-space-activities/space-situational (Accessed December 1 2014)

[3] "Space Situational Awareness", Space Safety & Sustainability Working Group, Space Generation Advisory Council, Austria, SSS Educational Series 2012
URL= http://www.agi.com/resources/educational-alliance-program/curriculum_exercises_labs/SGAC_Space%20Generation%20Advisory%20Council/space_situational_awareness.pdf (Accessed December 1 2014)

[4] Weeden, Brian. (Sept. 2014) "Space Situational Awareness Fact Sheet", Secure World Foundation. URL= http://swfound.org/media/1800/swf_ssa_fact_sheet_sept2014.pdf (Accessed December 1 2014)

[5] Weeden, Brian. (June 2011) "Space Situational Awareness: The Big Picture", 2011, Secure World Foundation, Presentation at European Space Surveillance Conference, Madrid Spain 7-9 URL= http://swfound.org/media/42072/SSA_The_Big_Picture-BW-2011.pdf (Accessed December 1 2014)

[6] United States Air Force, Air University, "Space Surveillance"
URL= http://www.au.af.mil/au/awc/awcgate/usspc-fs/space.htm (Accessed December 1 2014)

[7] Weeden, Brian., Cefola, Paul, and Sankaran J. (2010) "Global Space Situational Sensors," 2010 Advanced Maui Optical and Space Surveillance Conference, Maui, Hawaii, September 15-17.
URL= http://www.amostech.com/TechnicalPapers/2010/Integrating_Diverse_Data/Weeden.pdf
http://www.cissm.umd.edu/papers/display.php?id=541 (Accessed December 1 2014)

[8] Vallado, David A. and Griesbach, Jacob D. (2011) Simulating Space Surveillance Networks. Paper AAS 11-580 presented at the AAS/AIAA Astrodynamics Specialist Conference. July 31-August 4, Girdwood, AK.

[9] International Scientific Optical Network,
URL=http://lfvn.astronomer.ru/main/english.htm, http://lfvn.astronomer.ru/main/pulcoo.htm (Accessed December 1 2014)

[10] Captain Maskell, Paul., Oram, Lorne. (2008) "Sapphire: Canada's Answer to Space-Based Surveillance of Orbital Objects", URL= http://www.amostech.com/technicalpapers/2008/ssa_and_ssa_architecture/maskell.pdf (Accessed December 1 2014)

[11] Space Data Association, http://www.space-data.org/sda/ (Accessed December 1 2014)

[12] Staab S., Studer R (Ed.)(2009) Handbook on Ontologies, International Handbooks on Information Systems, Springer 2nd ed.

[13] Gomez-Perez, A., Fernandez-Lopez, M., Corcho, O. (2003) Ontological Engineering. Springer.

[14] Common Logic, http://www.iso.org/iso/catalogue_detail.htm?csnumber=39175

[15] Web Ontology Language URL= http://www.w3.org/TR/owl-features/, http://www.w3.org/TR/owl2-overview/ (Accessed December 1 2014)

[16] European Space Agency, "About SSA" URL= http://www.esa.int/Our_Activities/Operations/Space_Situational_Awareness/About_SSA (Accessed December 1 2014)

[17] "A Space Surveillance Ontology Captured in an XML Schema", October 2000, Mary K. Pulvermacher, Daniel L. Brandsma, John R. Wilson, MITRE, Center for Air Force C2 Systems, Bedford, Massachusetts.

[18] National Aeronautical and Space Administration (NASA). NASA Semantic Web for Earth and Environmental Terminology URL= https://sweet.jpl.nasa.gov/. (Accessed Feb 24 2016)

[19] International Virtual Observatory Alliance (IVOA). Ontology of Astronomical Object Types. URL= http://www.ivoa.net/documents/Notes/AstrObjectOntology/

[20] Hubner, S., Neumann, H., Stuckenschmidt, H., Schuster, G., Vogele, T., Visser, U., Wache, H. (2001) "Ontology-Based Integration of Information-A survey of Existing Approaches.". In Proceedings of the IJCAI-01 Workshop on Ontologies and Information Sharing, Seattle,

USA, August 4-5, 2001. P.108-118. URL=http://ftp.informatik.rwth-aachen.de/Publications/CEUR-WS/Vol-47/wache.pdf (Accessed Feb 2 2016)

[21] Pinto, H.S., Martins, J.P. (2001) "Ontology Integration: How to perform the Process" In Proceedings of the IJCAI-01 Workshop on Ontologies and Information Sharing, Seattle, USA, August 4-5, 2001. P.71-80. URL= http://ftp.informatik.rwth-aachen.de/Publications/CEUR-WS/Vol-47/pinto.pdf (Accessed Feb 2 2016)